\documentclass[10pt,twocolumn,letterpaper]{article}

\usepackage{cvpr}
\usepackage{times}
\usepackage{epsfig}
\usepackage{graphicx}
\usepackage{amsmath}
\usepackage{amssymb}
\usepackage{xspace}         %
\usepackage{mathtools}
\usepackage{makecell, verbatim, sidecap}

\usepackage[margin=4pt,font=small,labelfont=bf,labelsep=endash,tableposition=top]{caption}
\renewcommand{\texttt}[1]{${{\tt #1}}$}

\usepackage[pagebackref=true,breaklinks=true,letterpaper=true,colorlinks,bookmarks=false]{hyperref}

\cvprfinalcopy %

\ifcvprfinal\fi

\def\eg{{\it e.g.}\xspace}
\def\ie{{\it i.e.}\xspace}

\def\R{{\mathbb R}}

\renewcommand{\vec}[1]{\ensuremath{\pmb{#1}}}
\newcommand{\mat}[1]{\ensuremath{\mathbf{#1}}}
\newcommand{\set}[1]{\ensuremath{\mathscr{#1}}}

\makeatletter
\@tfor\next:=abcdefghijklmnopqrstuvwxyxz\do
{\begingroup\edef\x{\endgroup
		\noexpand\@namedef{v\next}{\noexpand\vec{\next}}%
	}\x}
\@tfor\next:=ABCDEFGHIJKLMNOPQRSTUVWXYZ\do
{\begingroup\edef\x{\endgroup
		\noexpand\@namedef{m\next}{\noexpand\mat{\next}}%
	}\x}
\@tfor\next:=ABCDEFGHIJKLMNOPQRSTUVWXYZ\do
{\begingroup\edef\x{\endgroup
		\noexpand\@namedef{s\next}{\noexpand\set{\next}}%
	}\x}
\makeatother

\begin{document}

\title{DirectPose: Direct End-to-End Multi-Person Pose Estimation}

\author{Zhi Tian, Hao Chen, Chunhua Shen\thanks{The first two authors
equally contributed to this work. Corresponding author: \texttt{chunhua.shen@adelaide.edu.au} }
\\[0.051cm]
The University of Adelaide, Australia
}

\maketitle
\begin{abstract}
We propose the first direct end-to-end multi-person pose estimation framework, termed DirectPose. Inspired by recent anchor-free object detectors, which directly regress the two corners of target bounding-boxes, the proposed framework directly predicts instance-aware keypoints for all the instances from a raw input image, eliminating the need for heuristic grouping in bottom-up methods or bounding-box detection and RoI operations in top-down ones. We also propose a novel Keypoint Alignment (KPAlign) mechanism, which overcomes the main difficulty---%
the feature mis-alignment between the convolutional features and predictions in this end-to-end framework. KPAlign improves the framework's performance by a large margin while still keeping the framework end-to-end trainable.  With the only post-processing non-maximum suppression (NMS), our proposed framework can detect multi-person keypoints with or without bounding-boxes in a single shot. Experiments demonstrate that the end-to-end paradigm can achieve competitive or better performance than previous strong baselines of both bottom-up and top-down methods. We hope that our end-to-end approach can provide a new perspective for the human pose estimation task.
\end{abstract}

\section{Introduction}
\begin{figure}[t]
\centering
\includegraphics[width=0.9\linewidth]{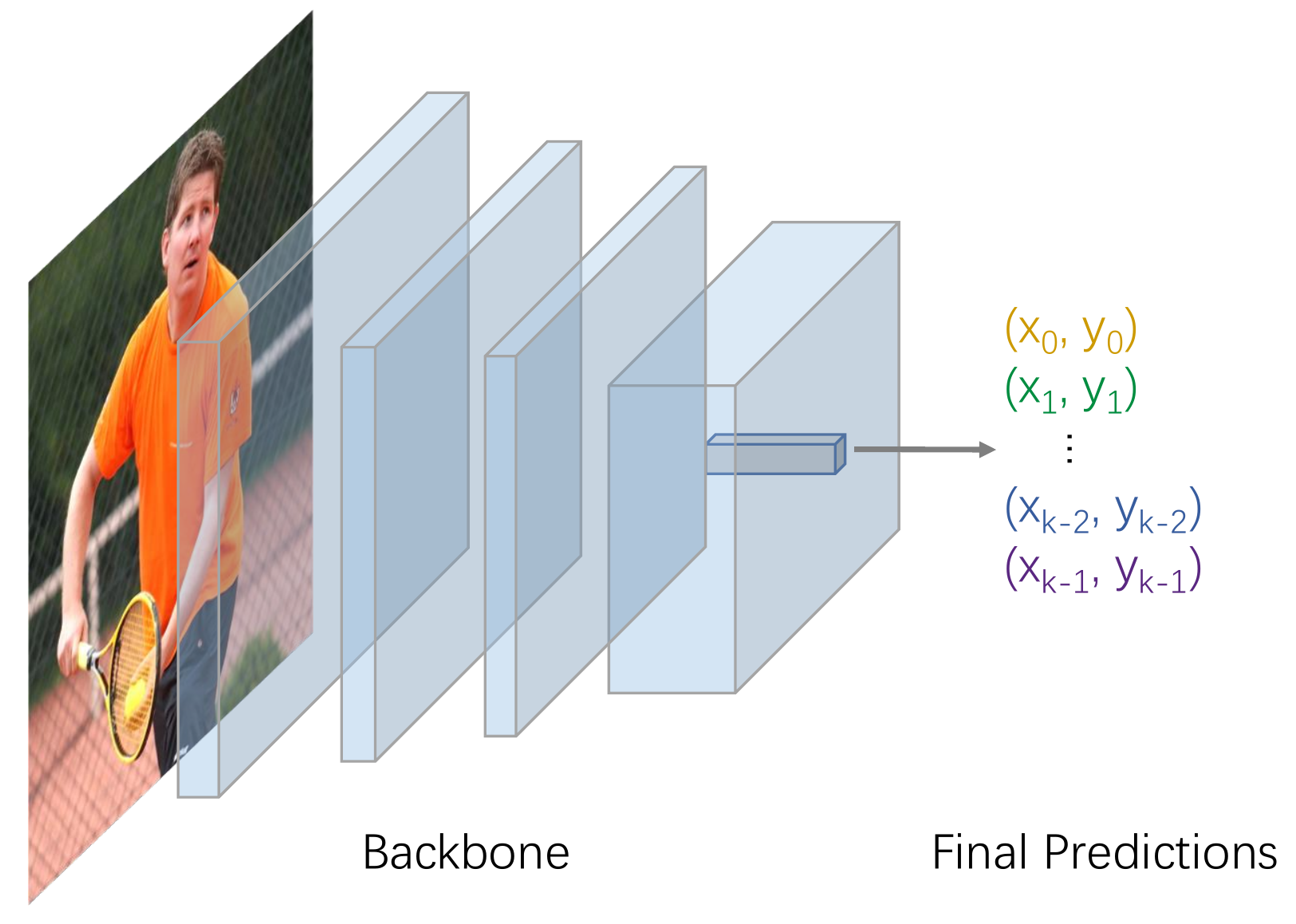}
\caption{The naive direct end-to-end keypoint detection framework. As shown in the figure,
the framework requires a single feature vector on the final feature maps to encode all the essential information of an instance (\eg, the precise locations of some keypoints for the instance, denoted as $(x_0, y_0), ... (x_{k-1}, y_{k-1})$). As shown in experiments, this single feature vector faces difficulty in preserving sufficient information for challenging instance-level recognition tasks such as keypoint detection.
Here we propose a keypoint alignment (KPAlign) module to overcome the issue.}
\label{fig:naive_framework}
\end{figure}
\begin{figure*}[t!]
	\centering
		\includegraphics[width=0.9\linewidth]{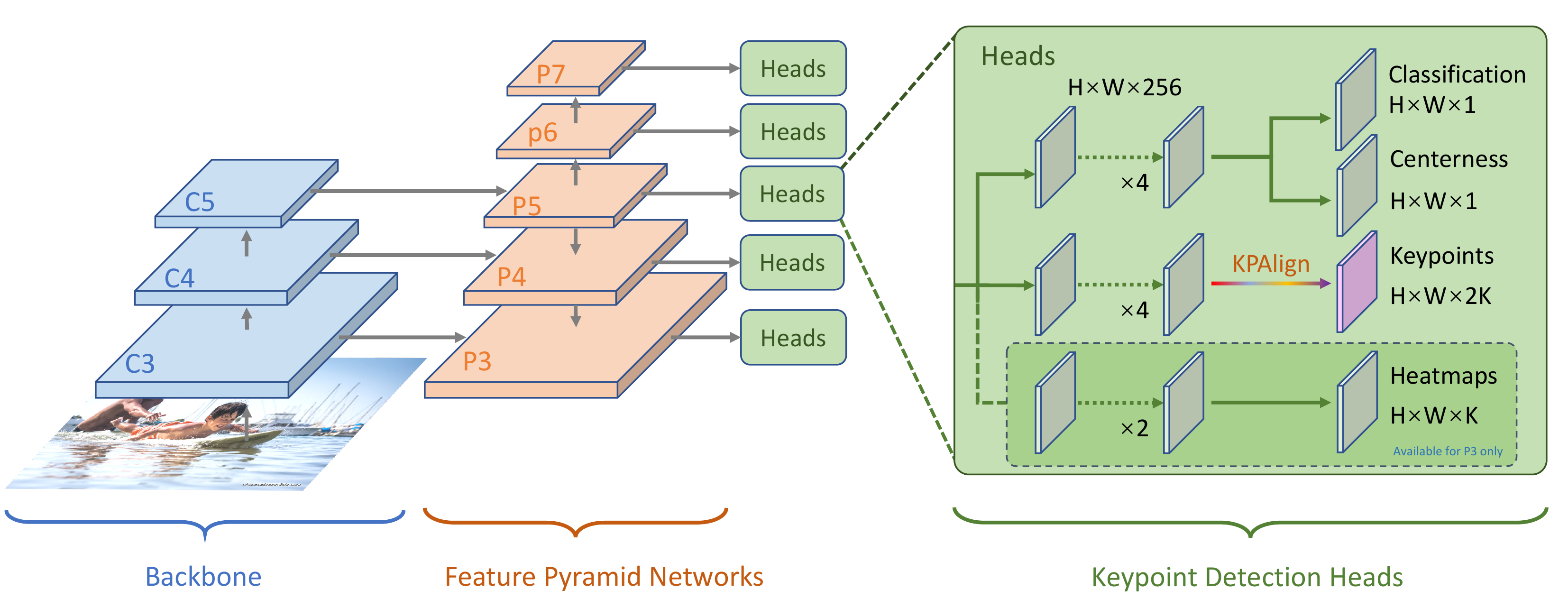}
	\caption{The proposed direct end-to-end multi-person pose estimation framework. The framework shares a similar architecture with one-stage object detectors such as FCOS \cite{tian2019fcos} but the bounding-box branch is replaced with a keypoint branch. KPAlign: the proposed keypoint alignment module, as described in Sec.~\ref{sec:kpalign}. Heatmaps: the branch for jointly heatmap-based learning and will be removed when testing. Keypoints: the branch for keypoint detection. Classification is from FCOS and used to classify the locations on the feature maps into ``person" or ``not person". Center-ness is also from FCOS and predicts how far the current location is from the center of its target object.}
	\label{fig:key_point_framework}
\end{figure*}

Multi-person pose estimation (a.k.a.\ keypoint detection) is a crucial step in the understanding of human behavior in images and videos. Previous methods for the task can be roughly categorized into bottom-up \cite{cao2017realtime, newell2017associative, papandreou2018personlab, pishchulin2016deepcut} and top-down \cite{he2017mask,sun2019deep,fang2017rmpe,chen2018cascaded} methods. Bottom-up methods first detect all the possible keypoints in an input image in an instance-agnostic fashion, which are followed by a grouping or assembling process to produce the final instance-aware keypoints. The grouping process is often heuristic and many tricks are involved to achieve a good performance. In contrast, top-down methods first detect each individual instance with a bounding-box and then reduce the task to single-instance keypoint detection. Although top-down methods can avoid the heuristic grouping process, they come with the price of long computational time since they cannot fully leverage the sharing computation mechanism of convolutional neural networks (CNNs). Moreover, the running time of top-down methods depends on the number of instances in the image,
making
them unreliable in some instant applications such as autonomous vehicles. Importantly, \textit{both bottom-up and top-down methods are not end-to-end}\footnote{Here we mean
	`direct end-to-end'; \ie, the model is trained end-to-end
	with keypoint annotations solely during training, and for inference,
	the model is able to map an input to keypoints for each individual instance without box detection and
	grouping post-processing.}, which
is in conflict with
deep learning's philosophy of learning everything together.

Recently, anchor-free object detection \cite{tian2019fcos, huang2015densebox} is emerging and has demonstrated superior performance than previous anchor-based object detection. These anchor-free object detectors directly regress two corners of a target bounding-box, without using pre-defined anchor boxes. The straightforward and effective solution for object detection gives rise to a question: \textit{can keypoint detection be solved with this simple framework as well?} It is easy to see that the keypoints for an instance can be considered as a special bounding-box with more than two corner points, and thus the task could be solved by attaching more output heads to the object detection networks. This solution is intriguing since 1) it is end-to-end trainable (\ie, directly mapping a raw input image to the desired instance-aware keypoints). 2) It can avoid the shortcomings of both top-down and bottom-up methods as it needs neither grouping or bounding-box detection. 3) It can unify object detection and keypoint detection in a single simple and elegant framework.

However, we show that such a naive approach performs
unsatisfactorily, mainly
due to the fact
that these object detectors resort to a single feature vector to regress all the keypoints of interest for an instance, with the hope  that the single feature vector can faithfully preserve the essential information (\eg, the precise locations of all the keypoints) in its receptive field, as shown in Fig.~\ref{fig:naive_framework}. While the single feature vector %
may be sufficiently good
to carry information for simple bounding-box detection as shown in %
\cite{tian2019fcos}, where only two corner points are involved in a bounding-box, it has difficulties in encoding
rich
information for the more challenging keypoint detection. As shown in our experiments, %
this straightforward
approach yields inferior performance.

In this work, we propose a keypoint alignment (KPAlign) mechanism to largely overcome the aforementioned problem of the solution. Instead of using a single feature vector to regress all the keypoints for an instance, the proposed KPAlign aligns the convolutional features with a target keypoint (or a group of keypoints) as possible as it can, and then predicts the location of the target keypoint(s) with the aligned features. Since the target keypoints and the used features are roughly aligned, the features are only required to encode the information in its neighborhood. It is evident that encoding the neighborhood is much easier than encoding the whole receptive field, which thus results in an improved performance. Moreover, the KPAlign module is differentiable, thus keeping the model end-to-end trainable. Additionally, it is well-known that learning a regression-based model is difficult. However, in this work, we find the regression task can largely benefit from a heatmap-based learning. As a result, we propose to jointly learn the two tasks during training. When testing, the heatmap-based branch is disabled and thus does not impose any overheads to the framework.

To summarize, the proposed one-stage regression-based keypoint detection enjoys the followings advantages over previous top-down or bottom-up approaches.
\begin{itemize}
\itemsep -.05112cm
\item The proposed framework is direct, totally
end-to-end trainable. To predict, it maps an input image to keypoints for each individual instance \textit{directly},
relying on neither intermediate operators like RoI feature cropping,
nor grouping post-processing, which sets our work apart from previous frameworks \cite{he2017mask, cao2017realtime} with multiple steps.
\item Our proposed framework can bypass the major shortcomings of both top-down and bottom-up methods. For example, compared to top-down methods, our framework can avoid the issue of early commitment and decouple computational complexity from the number of instances in an input image. Compared to bottom-up methods, our framework eliminates the heuristic post-processing assembling the detected keypoints into full-body instances.
\item Moreover, unlike previous top-down and bottom-up methods, both of which require a heatmap-based FCNs to detect keypoints and thus are with quantization error, our proposed framework directly regresses the precise coordinates of keypoints and thus decouple the output resolution of the networks and the precision of keypoint localization. It makes our framework have the potential to detect very dense keypoints (\ie, the keypoints crowd together).
\item Finally, the framework suggests that the keypoint detection task can also be solved with the same methodology as bounding-box detection (\ie, directly regressing all the keypoints or the corners of bounding-boxes), resulting in a unifying framework for both tasks.
\end{itemize}

\subsection{Related Work}
\paragraph{Top-down Methods:} Top-down methods \cite{sun2019deep, fang2017rmpe, pishchulin2012articulated, gkioxari2014using, papandreou2017towards, chen2018cascaded, xiao2018simple, ICCV2017Chen} break the multi-person pose estimation task into two sub-tasks -- person detection and single-person pose estimation. The person detection predicts a bounding-box for each instance in the input image. Next, the instance is cropped from the original image and a single-person pose estimation is applied to predict the keypoints for the cropped instance. Moreover, some approaches such as Mask R-CNN \cite{he2017mask} crop convolutional features rather than raw images, improving the efficiency of these methods. Top-down methods often have better performance but have higher computational complexity as it needs to repeatedly run the single-person pose estimation for each instance. Moreover, it also suffers from early commitment. In other words, it is difficult for these methods to recover an instance if it is missing in detection results.
\paragraph{Bottom-up Methods:} In contrast to top-down methods, which first identify individual instances by a detector, bottom-up methods \cite{cao2017realtime, pishchulin2016deepcut, insafutdinov2016deepercut} first detect all possible keypoints in an instance-agnostic fashion. Afterwards, a grouping process is employed to assemble these keypoints into full-body keypoints. Bottom-up methods can take advantage of the sharing convolutional computation, thus being faster than top-down methods. However, the grouping process is heuristic and involves many tricks and hyper-parameters. Recently, a one-stage framework \cite{Nie_2019_ICCV} makes the grouping process simpler. Compared to this work, our end-to-end framework further reduces the design complexity of a human pose estimation framework by directly mapping an input image to the desired keypoints.

Additionally, both top-down and bottom-up methods requires multiple steps to obtain the final keypoint detection results. Some of the steps are non-differentiable and make these methods impossible to be trained in an end-to-end fashion, which is the major difference between our methods and previous ones.
\section{Our Approach}
\begin{figure*}[t!]
	\centering
		\includegraphics[width=0.95\linewidth]{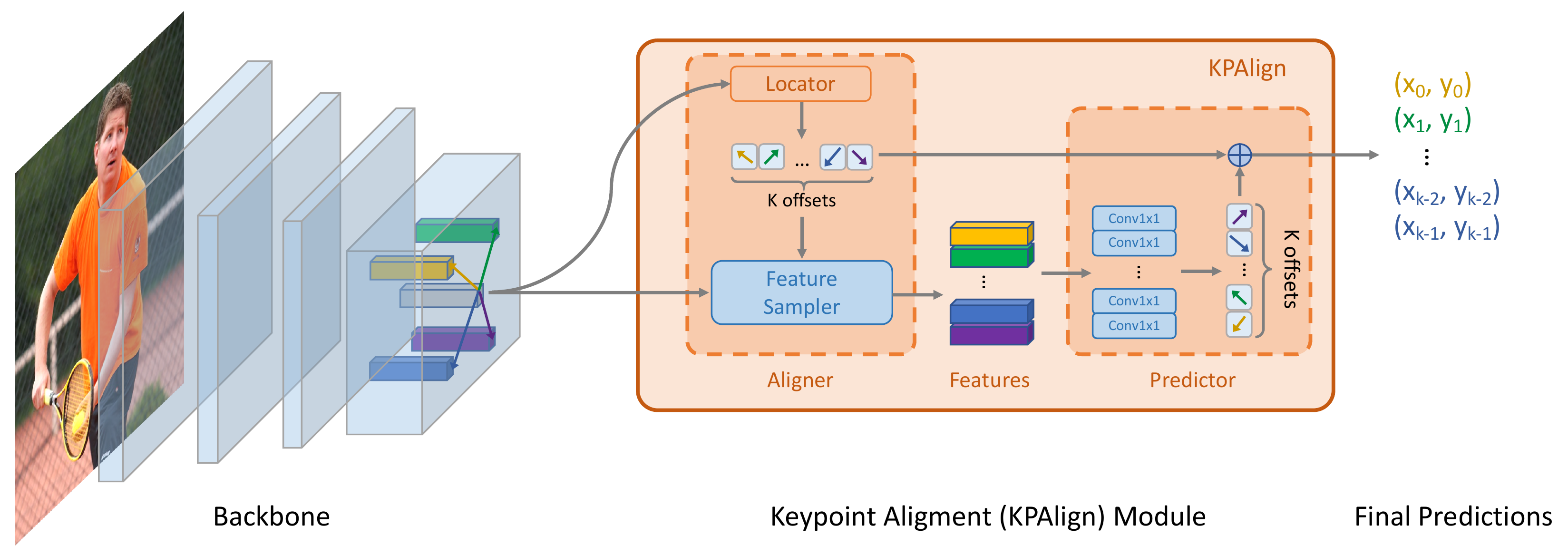}
	\caption{The proposed keypoint detection framework with the Keypoint Alignment (KPAlign) module. Feature pyramid networks (FPNs) are not shown here. The aligner consists a locator and a sampler. The locator is essentially a $3 \times 3$ convolution layer and predicts the rough locations of the keypoints. Next, the feature sampler samples feature vectors at these locations. %
	Thus, the aligner can roughly align the features and the predicted keypoints. The predictor employs these aligned feature vectors to make the final keypoint predictions.} \label{fig:kpalign}
\end{figure*}

Conceptually, our proposed keypoint detection framework is a simple extension of the anchor-free object detector FCOS \cite{tian2019fcos}, with one new output branch for keypoint detection. In this section, we first introduce FCOS detector and show how it can be extended to keypoint detection. Next, we illustrate our proposed KPAlign module, which allows the framework to leverage the feature-prediction alignment and improves the performance by a large margin. Finally, we present that how the jointly learning of the regression-based task and a heatmap-based task can be used to further boost the precision of keypoint localization.

\subsection{End-to-End Multi-Person Pose Estimation}
\paragraph{FCOS Detector:} FCOS detector is a recently-proposed anchor-free object detector. Unlike previous detectors such as RetinaNet \cite{lin2017focal} or Faster R-CNN \cite{ren2015faster}, FCOS eliminates anchor boxes and directly regresses the target bounding-boxes. It has been shown that FCOS can achieve even better performance than its anchor-based counterparts such as RetinaNet. To be specific, anchor-based detectors regard anchor boxes as training samples, which can be viewed as sliding windows on the input image. In contrast, FCOS views the pixels on the input image (or the locations on feature maps) as training samples, analogue to semantic segmentation. The pixels in a ground-truth box are viewed as positive samples and are required to regress the four offsets from the pixel (or location) to four boundaries of the ground-truth box (or equivalently the bounding-box's left-top and right-bottom coordinates relative to the pixel). Otherwise, the pixels are negative samples. The archiecture of FCOS is shown in Fig.~\ref{fig:key_point_framework} (changing the keypoint branch to a bounding-box branch and removing the heatmap prediction branch). FCOS shares a similar architecture with RetinaNet but has $\sim9\times$ less outputs.
\paragraph{Keypoint Representation:} It is straightforward to extend the bounding-box representation of FCOS to a keypoint representation. Specfically, we increase the scalars that each pixel regresses from $4$ to $2K$, where $K$ is the number of keypoints for each instance. Similarly, the $2K$ scalars denote the relative coordinates to the current pixel. In other words, we regard keypoints as a special bounding-box with $K$ corner points.
\paragraph{Our End-to-End Framework:} The keypoint representation results in our naive keypoint detection archiecture. As shown in Fig.~\ref{fig:key_point_framework}, it is implemented by applying a convolutional branch on all levels of the output feature maps of FPN \cite{lin2017feature} (\ie, $P_3, P_4, P_5, P_6$ and $P_7$). The downsampling ratios of these feature maps to the input image are $8$, $16$, $32$, $64$ and $128$, respectively. Note that the parameters of the branch are shared between FPN levels as in the bounding-box detection branch of the FCOS detector. The output channels of the branch is $2K$, where $K$ is the number of keypoints for each instance. The original bounding-box branch can be kept for simultaneous keypoint and bounding-box detection. Moreover, for keypoint-only detection, it is worth noting that we only use keypoint annotations without bounding-boxes. However, during training, FCOS requires a bounding-box for each instance to determine a positive or negative label for each location on the FPN feature maps. Here, we employ the minimum enclosing rectangles of keypoints of the instances as pseudo-boxes for computing the training labels.
\subsection{Keypoint Alignment (KPAlign) Module}\label{sec:kpalign}
We conduct preliminary experiments with the aforementioned naive keypoint detection framework. However, as shown in our experiments, it has inferior performance. We attribute the inferior performance to the lack of the alignment between the features and the predicted keypoints. Essentially, the naive framework makes use of a single feature vector at a location on the input feature maps to regress all the keypoints for an instance. As a result, the single feature vector is required to encode all the required information for the instance. This is difficult because many keypoints are far away from the center of the feature vector's receptive field and it has been shown in \cite{luo2016understanding} that the intensity of the feature's response decays quickly as the input signal deviates from the center of its receptive field. As shown in many FCN-based frameworks \cite{he2017mask, long2015fully}, keeping the feature and prediction aligned is crucial to good performance.
Thus
the feature only needs to encode the information in a local patch, which is much easier.

In this work, we propose a keypoint alignment (KPAlign) module to recover the feature-prediction alignment in the framework. KPAlign is used to replace the convolutional layer for the final keypoint detection in the naive framework and take as input the same feature maps, denoted as $\mF \in \R^{H \times W \times C}$, where $C$ being $256$ is the number of channels of the feature maps. Analogous to a convolution operation, KPAlign is densely slid through the input feature maps $\mF$. For simplicity, we take as an example a specific location $(i, j)$ on $\mF$ to illustrate how KPAlign works. As shown in Fig.~\ref{fig:kpalign}, KPAlign consists of two components --- an aligner $\zeta$ and a predictor $\phi$. The aligner consists of a locator and a feature sampler, and outputs the aligned feature vectors. The aligner can be formulated as,
\begin{equation}
\vo_0, \vo_1, ..., \vo_{K-1}, \vv_0, \vv_1, ..., \vv_{K-1} = \zeta(\mF),
\end{equation}
where $\vo_t \in \R^2$, produced by the locator in Fig.~\ref{fig:kpalign}, is the location where the feature vector used to predict the $t$-th keypoint of an instance should be sampled. $\vv_t \in \R^C$ is the sampled feature vector. Note that the location $\vo_{t}$ is defined over $\R^2$ and thus it can be fractional. Following \cite{dai2017deformable, he2017mask}, we make use of bilinear interpolation to compute the features at a fractional location. Additionally, the location is encoded as the coordinates relative to $(i, j)$ and thus is translation invariant.

Next, the predictor $\phi$ takes the outputs of the aligner as inputs to predict the final coordinates of the keypoints. As shown in Fig.~\ref{fig:kpalign}, the predictor includes $K$ convolution layers (\ie, one for each keypoint). Let us assume that we are looking for the $t$-th keypoint for the instance and let $\phi_t$ denote the $t$-th convlutional layer in the predictor. $\phi_t$ takes $\vv_t$ as input and predicts the coordinates of the $t$-th keypoint relative to the location where $\vv_t$ is sampled (\ie, $\vo_t$). Finally, the coordinates of the $t$-th keypoint, denoted as $\vx_t$, are the sum of the two sets of coordinates. Formally,
\begin{equation}
	\vx_t = \phi_t(\vv_t) + \vo_t, \qquad t = 0, 1, ..., K-1.
\end{equation}
Note that the coordinates need to be re-scaled by the down-sampling ratio of $\mF$. We omit the re-scaling operator here for simplicity. Note that all the operations in KPAlign module are differentiable and therefore {\it the whole model can be trained in an end-to-end fashion with standard back-propagation}, which sets our work apart from previous bottom-up or top-down keypoint detection frameworks such as CMU-Pose \cite{cao2017realtime} or Mask R-CNN \cite{he2017mask}. \emph{Being end-to-end trainable also makes the locator
be able to
learn to localize the keypoints without explicit supervision, which is
critically
important to KPAlign.
}

\paragraph{Grouped KPAlign:} The aforementioned KPAlign module is required to sample $K$ feature vectors for $K$ keypoints. This is actually not necessary because some keypoints (\eg, nose, eyes and ears) always populate in a local area. Therefore, we propose to group the keypoints and the keypoints in the same group will use the same feature vector, which reduces the number of sampled feature vectors from $K$ to $G$ and achieves a similar performance, where $G$ is the number of groups.
\paragraph{Using Separate Convolutional Features:} In the KPAlign described before, all of the keypoint groups use the feature maps $\mF$ as the input. However, we find that the performance can be improved \emph{remarkably} if we use separate feature maps for the $G$ keypoint groups (\ie, using $\mF_t, t = 0, 1, ..., G - 1$). In that way, the demand for the information encoded in a single $\mF_t$ can be further mitigated. In order to reduce the computational complexity, the number of channels of each $\mF_t$ is set as $\frac{C}{4}$ (\ie, from $256$ to $64$).
\paragraph{Where to Sample Features?} For the sake of convenience, the sampler in the aforementioned aligner samples features on the input feature maps of the locator, and therefore the predictor and locator take as inputs the same feature maps. However, it is not reasonable as the locator and predictor require different levels of feature maps. The locator predicts the initial but imprecise locations for all the keypoints (or keypoint groups) of an instance and thus requires high-level features with a larger receptive field. In contrast, the predictor needs to make precise predictions but only for the keypoints in a local area because the features have been aligned by the aligner. As a result, the predictor prefers high-resolution low-level features with a smaller receptive field. To this end, we feed lower levels of feature maps into the sampler. Specifically, if a locator uses feature maps $P_L$ and $P_L$ is not the finest feature maps, the sampler will take $P_{L-1}$ as the input. If $P_L$ is already the finest feature maps, the sampler will still sample on it.

\subsection{Regularization from Heatmap Learning}
It is well-known that regression-based tasks are difficult to learn \cite{glorot2010understanding, sun2018integral} and have poor generalization. That is why almost all previous keypoint detection methods \cite{cao2017realtime, he2017mask, wei2016convolutional, sun2019deep, newell2017associative} are based on heatmap prediction, which cast keypoint detection to a pixel-to-pixel prediction task. However, this reformulation makes an end-to-end training infeasible since it involves the \emph{non-differentiable} transformation between the heatmap-based prediction and desired continuous keypoint coordinates. Therefore, an end-to-end keypoint detection framework has to be regression-based.

As a result, we need to seek a way that can make the regression-based task easier to learn and generalize. To this end, given the fact that heatmap-based learning is much easier, we use the heatmap-based prediction task as an auxiliary task.
Thus,
the heatmap-based task can serve as a hint for the regression-based task and thus can regularize the task. In our experiments, the jointly learning significantly boosts the performance of the regression-based task. Note that \emph{the heatmap-based task is only used as an auxiliary loss during training}. It is
removed when testing.

\paragraph{Heatmap Prediction:} As shown in Fig.~\ref{fig:key_point_framework}, the heatmap prediction task takes as input the FPN feature maps $P_3$ with downsampling ratio being $8$. Afterwards, two $3 \times 3$ conv layers with channel being $128$ are applied here, which are followed by another $3 \times 3$ conv layer with output channel being $K$ for the final heatmap prediction, where $K$ is the number of keypoints for each instance. Previous heatmap-based keypoint detection methods \cite{cao2017realtime} generate un-normalized Gaussian distribution centered at each keypoint and thus generate the heatmaps in a per-pixel regression fashion. In contrast, since our framework does not
rely
on the heatmap prediction when testing, we
perform
a per-pixel classification here for simplicity. Note that we make use of multiple binary classifiers (\ie, one-versus-all) and therefore the number of output channels is $K$ instead of $K + 1$.

\paragraph{Ground-truth Heatmaps and Loss Function:} The ground-truth heatmaps are generated as follows. On the heatmaps, if a location is the nearest location to a keypoint with type $t$, the classification label for the location is set as $t$, where $t\in\{1, 2, ..., K\}$. Otherwise, the label is $0$.\footnote{%
        Strictly speaking, the generated ground-truth is a set of binary labels, rather than the conventional real-valued
        heatmap. We slightly abuse the term here.
}
Finally, in order to overcome the imbalance between positive and negative samples, we use focal loss \cite{lin2017focal} as the loss function.

\section{Experiments}
Our experiments are conducted on human keypoint detection task of the large-scale benchmark COCO dataset \cite{lin2014microsoft}. The dataset contains more than $250K$ person instances with $17$ annotated keypoints. Following the common practice \cite{cao2017realtime, he2017mask}, we use the COCO  \texttt{trainval35k} split ($57K$ images) for training and \texttt{minival} split ($5K$ images) as validation for our ablation study. We report our main results on the \texttt{test}-\texttt{dev} split ($20K$ images). Unless specified, we only make use of the human keypoint annotations without bounding-boxes. The performance is computed with Average Percision (AP) based on Object Keypoint Similarity (OKS).
\paragraph{Implementation Details:} Unless specified, ResNet-50 \cite{he2016deep} is used as our backbone networks. We use two training schedules. The first is quick and used to train a fast prototype of our models in ablation experiments. Specifically, the models are trained with stochastic gradient descent (SGD) on 8 V100 GPUs for $25$ epochs with a mini-batch of 16 images. For the main results on \texttt{test}-\texttt{dev} split, we use a longer training schedule; the models are trained for $100$ epochs with a mini-batch of 32 images. We set the initial learning rate to $0.01$ and use a linear schedule $base\_lr \times (1 - \frac{iter}{max\_iter})$ to decay it. Weight decay and momentum are set as 0.0001 and 0.9, respectively. We initialize our backbone networks with the weights pre-trained on ImageNet \cite{deng2009imagenet}. For the newly added layers, we initialize them as in \cite{lin2017focal}. When training, the images are randomly resized and horizontally flipped with probability being $0.5$, and the images are also randomly cropped into $800 \times 800$ patches. When testing, we run inference on the whole image and the testing images are resized to have its shorter side being $800$ and their longer side less or equal to $1333$. If bounding-box detection is available, NMS is applied to the detected bounding-boxes. Otherwise, we do NMS on the minimum enclosing rectangles of keypoints of the instances. The NMS threshold is set as $0.5$ for all experiments.
\subsection{Ablation Experiments}
\subsubsection{Baseline: the naive end-to-end framework}
\begin{table}
	\small
	\begin{center}
	\begin{tabular}{ l | c c c c c}
		\hline
		& AP$^{kp}$ & AP$^{kp}_{50}$ & AP$^{kp}_{75}$ & AP$^{kp}_{M}$ & AP$^{kp}_{L}$ \\
		\hline\hline
		Baseline & 43.4 & 73.8 & 45.1 & 38.9 & 50.9 \\
		w/ KPAlign$^\dag$ & 43.0 & 74.2 & 43.9 & 39.0 & 49.6 \\
		w/ KPAlign & \textbf{50.5} & \textbf{77.6} & \textbf{54.9} & \textbf{44.4} & \textbf{60.0} \\
		\hline
	\end{tabular}
	\end{center}
	\caption{Ablation experiments on COCO \texttt{minival} for the proposed KPAlign module. Baseline: the naive keypoint detection framework, as shown in Fig. \ref{fig:naive_framework}. ``w/ KPAlign$^\dag$": using the KPAlign module in the naive framework but disabling the aligner in it.
	``w/ KPAlign":
	using the full-featured KPAlign module.}
	\label{table:naive_w_kpalign}
\end{table}
We first experiment with the naive end-to-end keypoint detection framework in Fig.~\ref{fig:naive_framework} by replacing the bounding-box head in FCOS with the keypoint detection head. Moreover, as described before, we use pseudo-boxes to compute the label for each location on FPN feature maps during training.
As shown in Table~\ref{table:naive_w_kpalign}, the naive framework can only obtain low performance ($43.4\%$ in AP$^{kp}$). As mentioned before, the low performance is due to the misalignment between the features and keypoint predictions. In the following experiments, we will show that our proposed KPAlign can overcome the issue.
\subsubsection{Keypoint alignment (KPAlign) module}
In this section, we equip the above naive framework with our proposed KPAlign module. As shown in Fig.~\ref{fig:key_point_framework}, KPAlign serves as the final prediction layer, which was a standard convolutional layer in the naive framework. As shown in Table~\ref{table:naive_w_kpalign}, KPAlign improves the keypoint detection performance by a large margin (more than $7$ points in AP$^{kp}$).
In order to demonstrate that the improvement is indeed due to the retained alignment between the features and keypoint predictions rather than other factors (\eg, slightly more network parameters), we conduct another experiment in which the aligner of KPAlign is disabled. In other words, the offsets predicted by the locator are ignored and thus all the keypoints of an instance are predicted with the same features as in the naive framework. As shown in Table~\ref{table:naive_w_kpalign}, without the aligner, the performance drops dramatically to $43.0\%$ in AP$^{kp}$, which is nearly the same as the performance of the naive framework. Therefore, it is safe to claim that the improvement is due to the retained alignment.

\subsubsection{Grouped KPAlign}
\begin{table}
	\small
	\begin{center}
	\begin{tabular}{ l | c c c c c}
		\hline
		& AP$^{kp}$ & AP$^{kp}_{50}$ & AP$^{kp}_{75}$ & AP$^{kp}_{M}$ & AP$^{kp}_{L}$ \\
		\hline\hline
		KPAlign & 50.5 & 77.6 & 54.9 & 44.4 & 60.0 \\
		+ Grouped & 50.6 & 77.5 & 55.4 & 44.3 & 60.2 \\
		+ Sep. features & 51.4 & 78.2 & 55.6 & 45.6 & 60.6  \\
		+ Better sampling & \textbf{52.2} & \textbf{78.3} & \textbf{56.6} & \textbf{46.3} & \textbf{61.7} \\
		\hline
	\end{tabular}
	\end{center}
	\caption{Ablation experiments on COCO \texttt{minival} for the design choices in KPAlign. ``+ Grouped": using Grouped KPAlign, which has slightly better performance than the original but largely reduce the number of sampled feature vectors, thus being faster. ``+ Sep. features": using separate (but slimmer) feature maps for different keypoint groups, which has similar computational complexity but improved performance. ``+ Better sampling": the predictor samples features on finer feature maps (\ie, from $P_L$ to $P_{L-1}$).}
	\label{table:better_kp_align}
\end{table}
As described before, it is not necessary to sample $K$ (\ie, $17$ on COCO) feature vectors (one feature vector per keypoint) as some keypoints are always together and thus can be predicted with the same feature vector. In this experiments, we divide the keypoints into 9 groups\footnote{These groups respectively include (nose, left eye, right eye, left ear, right ear), (left shoulder, ), (left elbow, left wrist), (right shoulder, ), (right elbow, right wrist), (left hip, ), (left knee, left ankle), (right hip, ) and (right knee, right ankle).}, which reduces the number of the sampled feature vectors from $17$ to $9$ and makes the module faster. As shown in Table~\ref{table:better_kp_align}, the Grouped KPAlign can achieve slightly better performance than the original KPAlign. Therefore, in the sequel, we will use the Grouped KPAlign for all the following experiments. We also attempted other ways forming the groups but they achieve a similar performance.

\subsubsection{Using separate convolutional features}
As described before, using separate feature maps for these keypoint groups can improve the performance. Here we conduct an experiment for this. As shown in Table~\ref{table:better_kp_align}, using separate feature maps can boost the performance by $0.8\%$ in AP$^{kp}$ (from $50.6\%$ to $51.4\%$). Note that the number of channels of these separate feature maps is reduced from $256$ to $64$, and thus the model has similar computational complexity to the original one. As a result, the model using separate feature maps achieves a better trade-off between speed and accuracy.

\subsubsection{Where to sample features in KPAlign?}
In this experiment, the sampler samples on finer feature maps, as described in Sec.~\ref{sec:kpalign}, since the locator requires low-resolution high-level feature maps with a larger receptive field while the predictor prefers high-resolution low-level feature maps. As shown in Table~\ref{table:better_kp_align} (``+ Better Sampling"), using the sampling strategy can improve the performance to $52.2\%$. Note that using the sampling strategy does not increase the computational complexity of the model. Moreover, the better sampling strategy improves the AP$^{kp}_{50}$ and AP$^{kp}_{75}$ by $0.1\%$ and $1.0\%$, respectively, which implies that the sampling strategy can result in more accurate keypoint predictions because the improvement mainly comes from the AP$^{kp}$s at higher thresholds.
\subsubsection{Regularization from heatmap learning}
\begin{table}
	\small
    \begin{center}
	\begin{tabular}{ l | c c c c c}
		\hline
		& AP$^{kp}$ & AP$^{kp}_{50}$ & AP$^{kp}_{75}$ & AP$^{kp}_{M}$ & AP$^{kp}_{L}$ \\
		\hline\hline
		Baseline & 52.2 & 78.3 & 56.6 & 46.3 & 61.7 \\
		w/ $16\times$ Heatmap & 57.7 & 82.8 & 63.1 & 51.8 & 66.9 \\
		w/ \space\space$8\times$ Heatmap & 58.0 & 82.5 & 63.3 & 52.7 & 66.6 \\
		\hline
		+ Longer sched. & \textbf{63.1} & \textbf{85.6} & \textbf{68.8} & \textbf{57.7} & \textbf{71.3} \\
		\hline
	\end{tabular}
	\end{center}
	\caption{Ablation experiments on COCO \texttt{minival} for DirectPose with heatmap prediction. Baseline: without the heatmap learning. ``$16\times$ Heatmaps": predicting the heatmaps with downsampling ratio being $16$. ``$8\times$ Heatmaps": predicting the heatmaps with downsampling ratio being $8$ (\ie, using $P_3$). ``+ Long sched.": increasing the number of training epochs from $25$ to $100$. As shown in the table, learning with heatmap prediction can largely improve the performance. Moreover, using the design choices of the heatmaps (\eg, the resolution) have a small impact on the final performance, which is one of the advantages of our framework over previous bottom-up methods.} \label{table:e2e_pose_w_heatmaps}
\end{table}
As shown in Table~\ref{table:e2e_pose_w_heatmaps} (``w/ $8\times$ Heatmaps"), by jointly learning the regression-based model with a heatmap prediction task, the performance of the regression-based task can be largely improved from $52.2\%$ to $58.0\%$. Note that the heatmap prediction is only used during training to provide the multi-task regularization. Moreover, we also conduct experiments with the heatmap prediction with a lower resolution (\ie, ``w/ $16\times$ Heatmaps"). As shown in Table~\ref{table:e2e_pose_w_heatmaps}, even with the low-resolution heatmaps, the model can still yield a similar performance. This suggests that our method is not sensitive to the design choices for the heatmap learning and thus eliminates the heuristic tuning for the heatmap branch. This sets our method apart from previous heatmap-based bottom-up methods, whose performance highly depends on the design of the heatmap branch (\eg, the heatmaps' resolution and etc.).

Moreover, we find that our method is highly under-fitting and previous methods such as \cite{sun2019deep} with heatmaps learning are trained with much more epochs than ours, and therefore we increase the number of epochs from $25$ to $100$. As shown in Table~\ref{table:e2e_pose_w_heatmaps}, this improves the performance by $5.1\%$ in AP$^{kp}$.
\subsection{Combining with Bounding Box Detection}
\begin{table}
	\small
	\begin{center}
	\begin{tabular}{ c | c c c | c c c}
		\hline
		w/ BBox & AP$^{bb}$ & AP$^{bb}_{50}$ & AP$^{bb}_{75}$ & AP$^{kp}$ & AP$^{kp}_{50}$ & AP$^{kp}_{75}$ \\
		\hline\hline
		 & - & - & - & \textbf{63.1} & \textbf{85.6} & \textbf{68.8} \\
		\checkmark & 55.3 & 81.5 & 59.9 & 61.5 & 84.3 & 67.5 \\
		\hline
	\end{tabular}
	\end{center}
	\caption{Our framework with person bounding-box detection on COCO \texttt{minival}. The proposed framework can achieve reasonable person detection results ($55.3\%$ in AP). As a reference, the Faster R-CNN person detector in Mask R-CNN \cite{he2017mask} achieves $53.7\%$ in AP.}
	\label{table:e2e_pose_w_bbox}
\end{table}
As mentioned before, by simply adding a bounding-box branch, the proposed framework can simultaneously detect bounding boxes and keypoints. Here we confirm it by the experiment. The bounding-box detection is implemented by adding the original box detection head of FCOS to the framework. As shown in Table~\ref{table:e2e_pose_w_bbox}, our framework can achieve a reasonable person detection performance, which is similar to the Faster R-CNN detector in Mask R-CNN ($55.3\%$ vs. $53.7\%$). Although Mask R-CNN can also simultaneously detect bounding-boxes and keypoints, we further unify the two tasks into the same methodology.
\subsection{Comparisons with State-of-the-art Methods}
\begin{table}
	\small
	\begin{center}
	\begin{tabular}{ l | c c c c c }
		\hline
		Method & AP$^{kp}$ & AP$^{kp}_{50}$ & AP$^{kp}_{75}$ & AP$^{kp}_{M}$ & AP$^{kp}_{L}$ \\
		\hline\hline
		\multicolumn{6}{c}{\textbf{Top-down Methods}} \\
		\hline
		Mask R-CNN \cite{he2017mask} & 62.7 & 87.0 & 68.4 & 57.4 & 71.1 \\
		CPN \cite{chen2018cascaded} & 72.1 & 91.4 & 80.0 & 68.7 & 77.2 \\
		RMPE \cite{fang2017rmpe} & 72.3 & 89.2 & 79.1 & 68.0 & 78.6 \\
		CFN \cite{huang2017coarse} & 72.6 & 86.1 & 69.7 & 78.3 & 64.1 \\
		HRNet-W48 \cite{sun2019deep} & \textbf{75.5} & \textbf{92.5} & \textbf{83.3} & \textbf{71.9} & \textbf{81.5} \\
		\hline
		\multicolumn{6}{c}{\textbf{Bottom-up Methods}} \\
		\hline
		CMU-Pose$^{*\dag}$ \cite{cao2017realtime} & 61.8 & 84.9 & 67.5 & 57.1 & 68.2 \\
		AE \cite{newell2017associative} & 56.6 & 81.8 & 61.8 & 49.8 & 67.0 \\
		AE$^*$ & 62.8 & 84.6 & 69.2 & 57.5 & 70.6 \\
		AE$^{*\dag}$ & 65.5 & 86.8 & 72.3 & 60.6 & 72.6 \\
		PersonLab \cite{papandreou2018personlab} & 65.5 & 87.1 & 71.4 & 61.3 & 71.5 \\
		PersonLab$^{\dag}$ & \textbf{67.8} & \textbf{89.0} & \textbf{75.4} & \textbf{64.1} & \textbf{75.5} \\
		\hline
		\multicolumn{6}{c}{\textbf{Direct End-to-end Methods}} \\
		\hline
		\textbf{Ours (R-50)} & 62.2 & 86.4 & 68.2 & 56.7 & 69.8 \\
		\textbf{Ours (R-50)$^{\dag}$} & 63.0 & 86.8 & 69.3 & 59.1 & 69.3 \\
		\textbf{Ours (R-101)} & 63.3 & 86.7 & 69.4 & 57.8 & 71.2 \\
		\textbf{Ours (R-101)$^{\dag}$} & \textbf{64.8} & \textbf{87.8} & \textbf{71.1} & \textbf{60.4} & \textbf{71.5} \\
		\hline
	\end{tabular}
	\end{center}
	\caption{The performance of our proposed end-to-end framework on COCO \texttt{test}-\texttt{dev} split. $^*$ and $^\dag$ respectively denote using refining and multi-scale testing. As shown in the table, the new end-to-end framework achieves competitive or better performance than previous strong baselines (\eg, Mask R-CNN and CMU-Pose).}
	\label{table:comparsions_with_sota}
\end{table}
In this section, we evaluate the proposed end-to-end keypoint detection framework on MS-COCO \texttt{test}-\texttt{dev} split and compare it with previous bottom-up and top-down ones. We make use of the best model in ablation experiments. As shown in Table~\ref{table:comparsions_with_sota}, without any bells and whistles (\eg, multi-scale and flipping testing, the refining in \cite{cao2017realtime, newell2017associative}, and any other tricks), the end-to-end framework achieves $62.2\%$ and $63.3\%$ in AP$^{kp}$ on COCO \texttt{test}-\texttt{dev} split, with ResNet-50 and ResNet-101 as the backbone, respectively. With multi-scale testing, our framework can achieve $63.0\%$ and $64.8\%$ with ResNet-50 and ResNet-101, respectively. Qualitative results will be provided in the supplemental material.

\paragraph{Compared to Bottom-up Methods:} The performance of our ResNet-50 based end-to-end framework is better ($62.2\%$ vs. $61.8\%$) than the strong baseline CMU-Pose \cite{cao2017realtime} that uses multi-scale testing and post-processing with CPM \cite{wei2016convolutional}, and filters the results with an object detector. Our framework also achieves much better performance than the bottom-up method AE \cite{newell2017associative} ($63.3\%$ vs. $56.6\%$) and is even better than the method with refining. Compared to PersonLab, with the same backbone ResNet-101 and single-scale testing, our proposed framework also has a competitive performance with it ($63.3\%$ vs. $65.5\%$). Note that our proposed framework is much simpler than these bottom-up methods, in both training and testing.

\paragraph{Compared to Top-down Methods:} With the same backbone ResNet-50, the proposed method has a similar performance with previous strong baseline Mask R-CNN ($62.2\%$ vs. $62.7\%$). Our model is still behind other top-down methods. However, it is worth noting that these methods often employ a separate bounding-box detector to obtain person instances. These instances are then cropped from the original image and a single person pose estimation method is separately applied to each the cropped image to obtain the final results. As noted before, this strategy is slow as it cannot take advantage of the sharing computation mechanism in CNNs. In contrast, our proposed end-to-end framework is much simpler and faster since it directly maps the raw input images to the final instance-aware keypoint detections with a fully convolutional network.
\paragraph{Timing:} The averaged inference time of our model on COCO \texttt{minival} split is 74ms and 87ms per image with ResNet-50 and ResNet-101, respectively, which is slightly faster than Mask R-CNN with the same hardware and backbones (Mask R-CNN takes 78ms per image with ResNet-50). Additionally, the running time of Mask R-CNN depends on the number of the instances while our model, similar to one-stage object detectors, has nearly constant inference time for any number of instances.

\section{More Discussions and Results}
Here
 1) we further compare our proposed DirectPose against the recent SPM method \cite{Nie_2019_ICCV}. 2) We show the visualization results of the proposed KPAlign module. 3) The loss curves of training with or without the proposed heatmap learning are shown as well. 4) We show some final detection results with or without the simultaneous bounding-box detection.

\subsection{Comparison to SPM}

Here we highlight the difference between our proposed DirectPose and SPM \cite{Nie_2019_ICCV}. SPM makes use of Hierarchical Structured Pose Representations (Hierarchical SPR) to avoid learning the long-range displacements between the root and the keypoints, which shares a similar motivation with the proposed KPAlign. However, SPM considers all the key nodes (including the root nodes and intermediate nodes) in the hierarchical SPR as the regression targets, and instance-agnostic heatmaps are used to predict these nodes. This is similar to  OpenPose \cite{cao2017realtime} with the only exception that SPM predicts these nodes instead of the final keypoints, and thus the predicted nodes are also instance-agnostic. As a result, SPM still needs a grouping post-processing to assemble the detected nodes into full-body poses. In contrast, the proposed KPAlign only requires the coordinates of the final keypoints as the supervision, and aligns the features and the predicted keypoints in an unsupervised fashion. Hence, our proposed framework can directly predict the desired instance-aware keypoints, without the need for any form of grouping post-processing.

\subsection{Visualization of KPAlign}
\begin{figure*}[t]
\begin{center}
  \includegraphics[width=\linewidth]{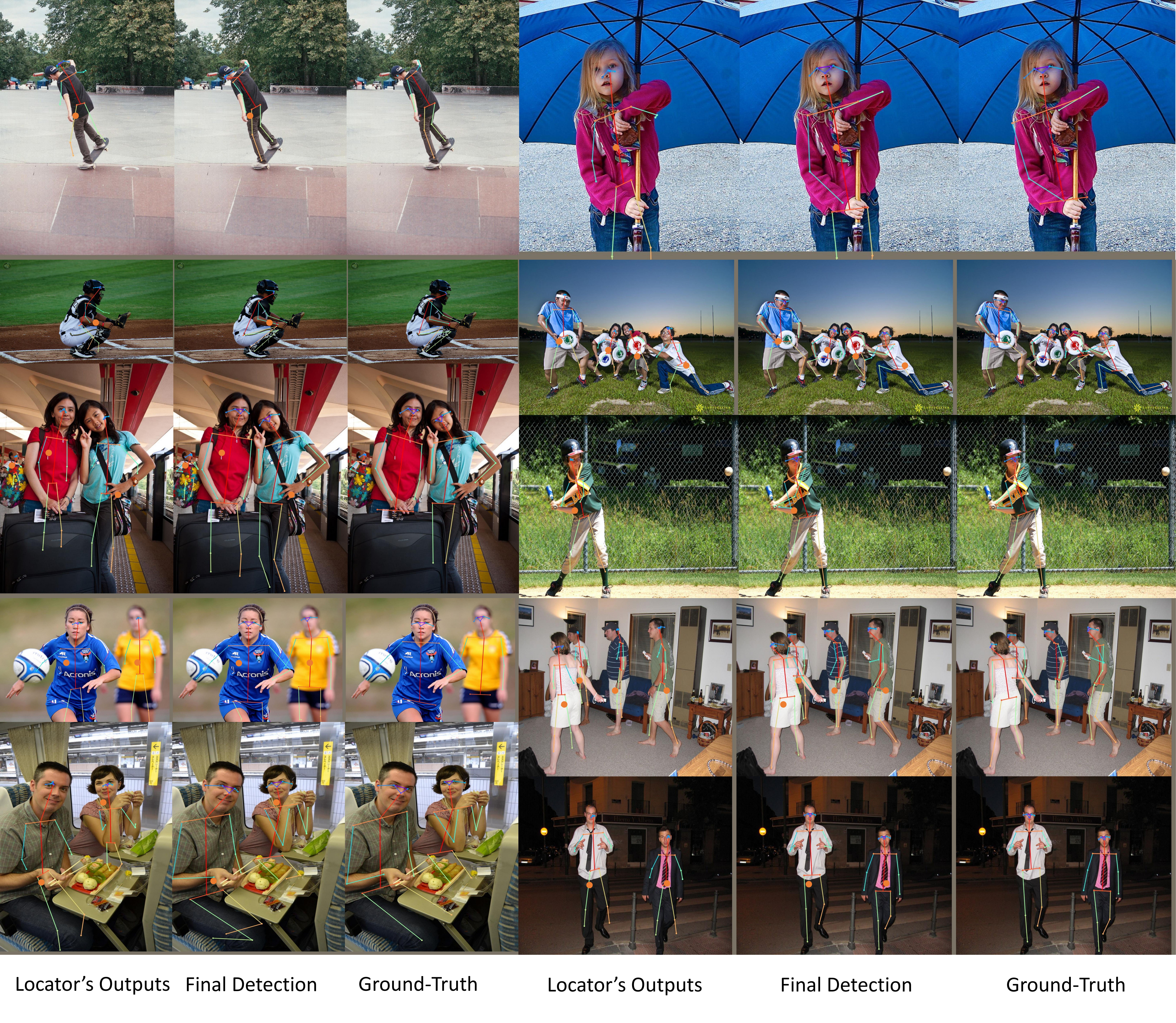}
\end{center}
  \caption{Visualization results of KPAlign on MS-COCO \texttt{minival}. The first image in each group shows the outputs of the locator in KPAlign (\ie, the locations where the sampler samples the features used to predict the keypoints). The orange point denotes the original location where the features will be used if KPAlign is not used. The second image shows the final keypoint detection results. As shown in the figure, the proposed KPAlign can make use of the features near the keypoints to predict them. The final image shows that the ground-truth keypoints. Zoom in for a better look.}
\label{fig:more_visualization_of_kpalign}
\end{figure*}

The visualization results of KPAlign are shown in Fig.~\ref{fig:more_visualization_of_kpalign}. As shown in the figure, the proposed KPAlign can make use of the features near the keypoints to predict them. Thus, the feature vectors can avoid encoding the keypoints far from their spatial location, which results in improved performance.

\subsection{Training Losses of using Heatmap Learning}
\begin{figure}[t!]
\begin{center}
  \includegraphics[width=.75\linewidth]{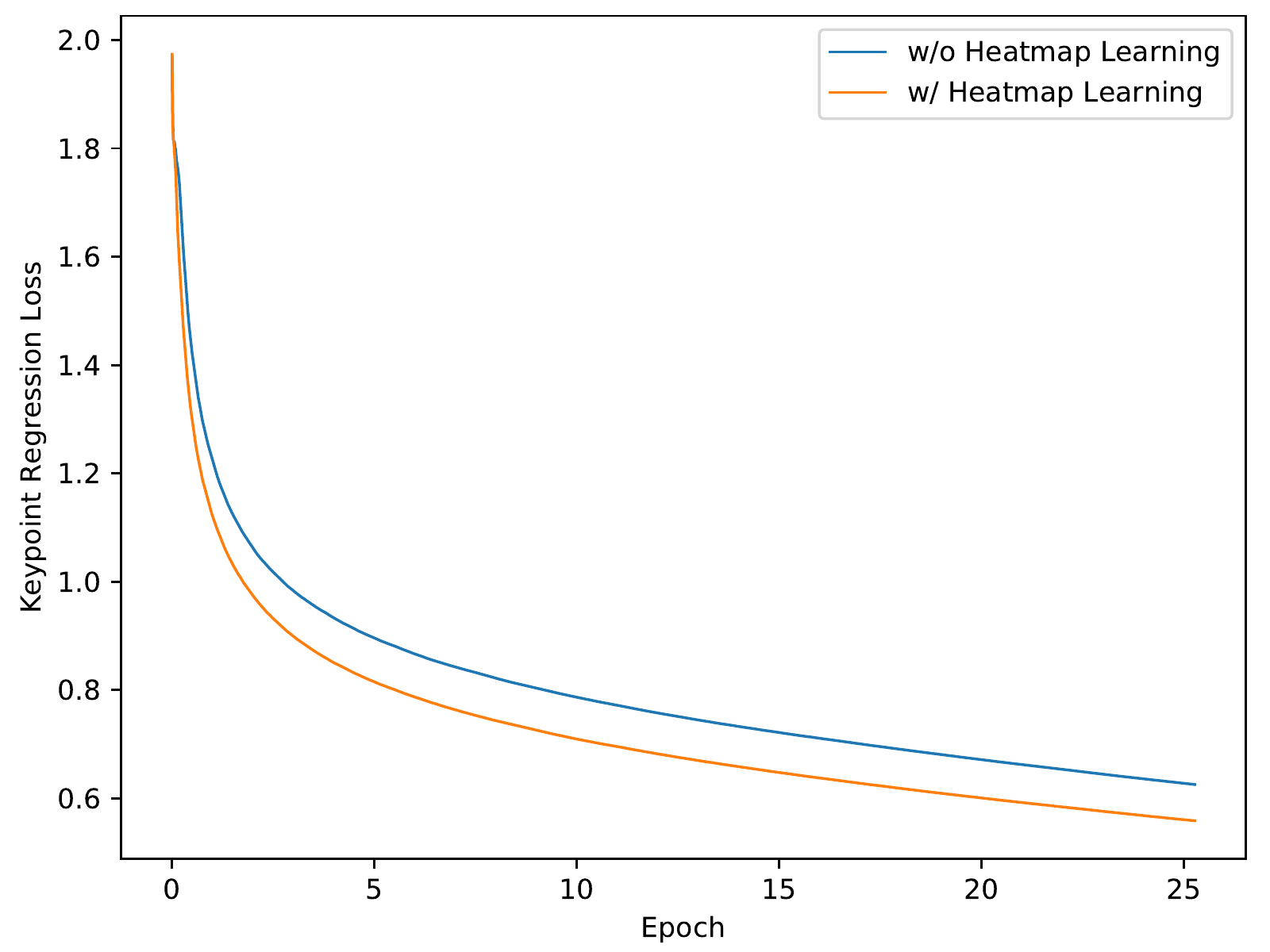}
\end{center}
  \caption{The loss curves of training with or without the heatmap learning. As shown in the figure, with the heatmap learning, the model can achieve a significantly lower loss value and thus much better performance.}
\label{fig:training_losses}
\end{figure}
In order to demonstrate the impact of the heatmap learning, we plot the loss curves of training with or without the heatmap learning in Fig.~\ref{fig:training_losses}. As shown in the figure, the heatmap learning can greatly help the training of the model and make the model achieve a much lower loss value, thus resulting in much better performance.

\subsection{Visualization of Keypoint Detections}
\begin{figure*}[t]
\begin{center}
  \includegraphics[width=\linewidth]{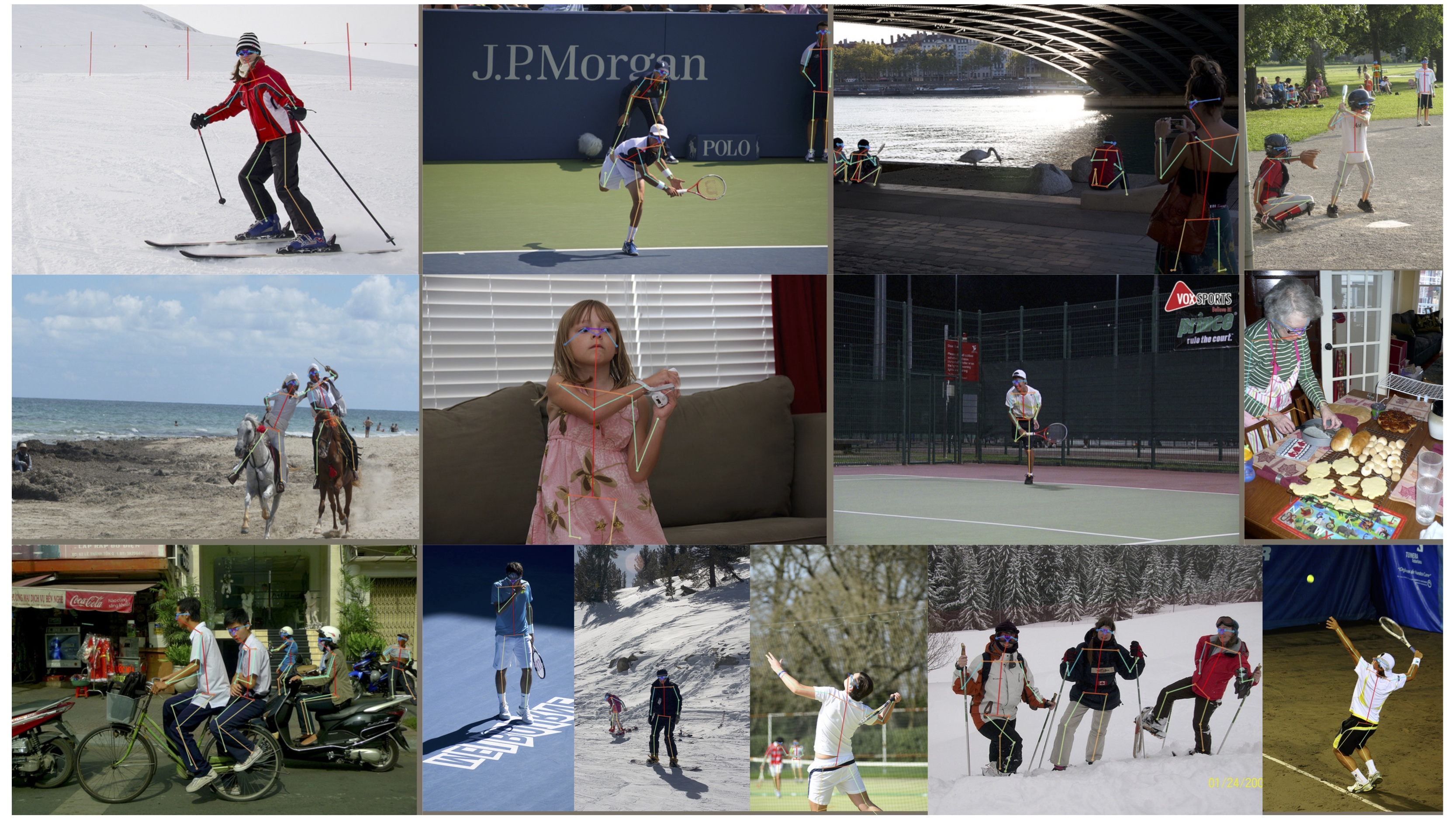}
\end{center}
  \caption{Visualization results of the proposed DirectPose on MS-COCO \texttt{minival}. DirectPose can directly detect a wide range of poses. Note that some small-scale people do not have ground-truth keypoint annotations in the training set of MS-COCO, thus they might be missing when testing.}
\label{fig:visualization_r101}
\end{figure*}

\begin{figure*}[t]
\begin{center}
  \includegraphics[width=\linewidth]{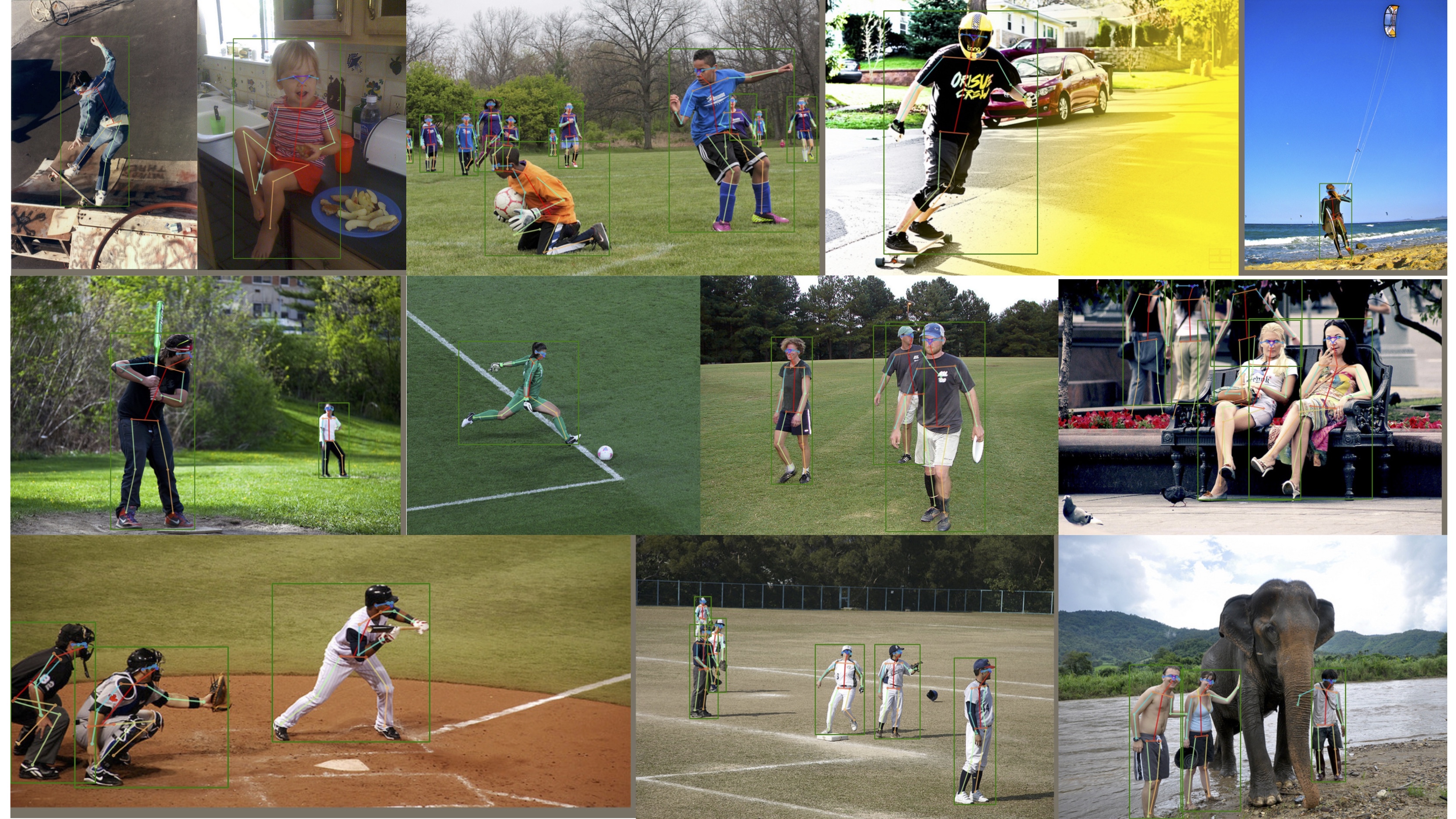}
\end{center}
  \caption{Visualization results of the proposed DirectPose with the simultaneous bounding-box detection on MS-COCO \texttt{minival}.}
\label{fig:visualization_r50_w_bbox}
\end{figure*}

We show more visualization results of DirectPose in Fig.~\ref{fig:visualization_r101}. As shown in the figure, the proposed DirectPose can directly detect all the desired instance-aware keypoints without the need for the grouping post-processing or bounding-box detection. The results of the proposed DirectPose with simultaneous bounding-box detection are also shown in Fig.~\ref{fig:visualization_r50_w_bbox}.

\section{Conclusion}
We have proposed the first direct end-to-end human pose estimation framework, termed DirectPose. Our proposed model is end-to-end trainable and can directly map a raw input image to the desired instance-aware keypoint detections within constant inference time, eliminating the need for the grouping post-processing in bottom-up methods or the bounding-box detection and RoI operations in top-down ones. We also proposed a keypoint alignment (KPAlign) module to overcome the major difficulty that is the lack of the alignment between the convolutional features and the predictions in the end-to-end model, significantly improving the keypoint detection performance. Additionally, we further improve the regression-based task's performance by jointly learning it with a heatmap-based task. Experiments demonstrate that the new end-to-end method can obtain competitive or better performance than previous bottom-up and top-down methods.

{\small
\bibliographystyle{ieee_fullname}
\bibliography{egbib}
}

\end{document}